\PassOptionsToPackage{sort&compress}{natbib}
\let\oldvec\vec
\documentclass[twocolumn,fleqn,natbib]{svjour2}

\let\vec\oldvec
\usepackage{amsmath}
\usepackage{graphicx}
\graphicspath{{./figures/}{./figures/plots/tex/}}
\usepackage{tabu} 
\usepackage{booktabs} 
\usepackage{adjustbox} 
\usepackage{array} 
\usepackage{multirow} 
\usepackage{microtype}
\usepackage[english]{babel}
\usepackage{color}
\usepackage{subfigure}
\usepackage{hyphenat}
\usepackage{flushend}
\usepackage{hyperref}

\setcitestyle{numbers,square}

\newcommand\w[1]{\makebox[\ww]{$#1$}}
\newcommand*\Figure[1]{Fig.~\ref{#1}}
\newcommand*\Table[1]{Table~\ref{#1}}
\newcommand*\Section[1]{Section \mbox{\nameref{#1}}}

\journalname{Journal of Real-Time Image Processing}

\title{Parallel Wavelet Schemes for Images}

\subtitle{How to make the wavelet transform friendly to parallel architectures}

\author{David Barina \and Michal Kula \and Pavel Zemcik}

\institute{
	Centre of Excellence IT4Innovations\\
	Faculty of Information Technology\\
	Brno University of Technology\\
	Bozetechova 1/2, 612 66 Brno\\
	Czech Republic\\
	\email{\{ibarina,ikula,zemcik\}@fit.vutbr.cz}
}

\date{Received: date / Revised: date}

\def\M{\mathrm{M}}
\def\N{\mathrm{N}}
\def\T{\mathrm{T}}
\def\S{\mathrm{S}}
\def\R{\mathrm{R}}
\def\x{\mathbf{x}}
\def\y{\mathbf{y}}
\def\L{\mathrm{L}}
\def\H{\mathrm{H}}
\def\LL{\mathrm{LL}}
\def\HL{\mathrm{HL}}
\def\LH{\mathrm{LH}}
\def\HH{\mathrm{HH}}
\def\P{\mathrm{P}}
\def\U{\mathrm{U}}
\def\V{\mathrm{V}}

\begin{document}

\maketitle

\begin{abstract}
\hspace{10pt}
In this paper, we introduce several new schemes for calculation of discrete wavelet transforms of images.
These schemes reduce the number of steps and, as a consequence, allow to reduce the number of synchronizations on parallel architectures.
As an additional useful property, the proposed schemes can reduce also the number of arithmetic operations.
The schemes are primarily demonstrated on CDF 5/3 and CDF 9/7 wavelets employed in JPEG 2000 image compression standard.
However, the presented method is general, and it can be applied on any wavelet transform.
As a result, our scheme requires only two memory barriers for \mbox{2-D} CDF 5/3 transform compared to four barriers in the original separable form or three barriers in the non-separable scheme recently published.
Our reasoning is supported by exhaustive experiments on high-end graphics cards.
\end{abstract}


\section{Introduction}
\label{sec:introduction}

The two-dimensional discrete wavelet transform (DWT) is a signal-processing transform suitable as a basis for sophisticated compression algorithms.
For example, JPEG 2000, an image coding system, is based on such compression technique.
This paper focuses on the Cohen--Daubechies--Feauveau (CDF) 5/3 and 9/7 wavelets \cite{Cohen1992}, which are often used for image compression.
However, the methods are general, and they are not limited to any specific type of transform.
Of course, plenty of other applications are built over the discrete wavelet transform.

The one-dimensional discrete wavelet transform has undergone a gradual development in the last few decades.
Probably, the most important advance is the discovery of a factoring algorithm \cite{Daubechies1998} referred to as the lifting scheme.
In this context, the discrete wavelet transform or two-band subband filtering can be represented by a polyphase matrix.
The lifting scheme algorithm decomposes any wavelet transform with finite filters into a finite sequence of lifting steps, while reducing the number of arithmetic operations.
The decomposition corresponds to a factorization of the polyphase matrix filters into elementary matrices.
The resulting coefficients of \mbox{1-D} transform are formed in two subbands.
The subbands correspond to low-pass (L) and high-pass (H) filtered subsampled variants of the original signal.

In case of two-dimensional transform \cite{Mallat1989}, one level of the transform can be realized using a separable decomposition scheme.
In this scheme, the coefficients are evaluated by successive horizontal and vertical \mbox{1-D} filtering, resulting in four disjoint groups (LL, HL, LH, and HH subbands).
A naive algorithm of \mbox{2-D} transform computation directly follows the horizontal and vertical filtering loops.
As a consequence, the number of elementary polyphase matrices is doubled.

Unfortunately, this separable computation does not reflect the requirements of the parallel architectures where the scheme will need twice as many synchronizations.
Such synchronizations often form a bottleneck of the overall calculation.
State-of-the-art algorithms fuse the horizontal and vertical loops into a single one, which results in the single-loop approach.
However, the number of the elementary polyphase matrices and thus the number of memory barriers remain unaffected.

To solve the outlined issue, we propose several novel spatial lifting structures computing the \mbox{2-D} discrete wavelet transform with reduced number of memory barriers.
These lifting structures are presented in the order in which they were gradually derived.
The presented work is accompanied by exhaustive performance experiments.

A typical representative of parallel architectures is the graphics processing unit (GPU) capable of executing a general-purpose program.
Actually, this is the architecture used to evaluate the performance of algorithms presented in this paper.
We have employed OpenCL language for writing underlying implementations.
These implementations were then subject of performance measurements on significant graphics cards of two biggest vendors.

In order to avoid misunderstandings, it should be noted that the schemes presented in this paper do not affect an image compression ratio nor quality.
The schemes only affect the speed in which the compression is completed.
Since practical applications require a multi-level discrete wavelet decomposition, the question of how to compute this multi-scale pyramid may arise.
In this case, the schemes discussed in this paper can simply be applied in a sequence exchanging intermediate results through a off-chip memory (a global memory in the case of GPU).
Another possibility is to apply this sequence on blocks exchanging the results using a fast on-chip memory (a local memory on GPU).
The latter possibility was used, e.g., in \cite{Matela2009,Arguello2012} employing the naive algorithm of \mbox{2-D} transform computation.

The rest of the paper is organized as follows.
\Section{sec:related-work} presents the theory in the necessary level of detail.
This theory includes the lifting scheme basics and the spatial lifting structures recently proposed.
Subsequent \Section{sec:proposed-schemes} derives the new spatial lifting structures.
Additionally, \Section{sec:improvements} presents a simple trick proposed in order to reduce the number of arithmetic operations.
\Section{sec:evaluation} and \Section{sec:performance} offer a thorough performance evaluation.
Finally, \Section{sec:conclusions} summarizes the paper.

\section{Related Work}
\label{sec:related-work}

In this paper, the well-known $z$-transform notation is employed for the description of FIR filters.
The transfer function of the FIR filter $h_{k}$ is a Laurent polynomial defined as
\begin{align}
	H(z) = \sum_{k=k_0}^{k_1-1} \, h_{k} \, z^{-k},
\end{align}
where $k_0$ denotes the smallest and $k_1-1$ denotes the largest integer number $k$ for which $h_{k}$ is non-zero.
The degree of a Laurent polynomial $H(z)$ is defined as $ |H(z)| = k_1-k_0-1 $.
Similarly, the transfer function of the two-dimensional FIR filter $h_{k_m,k_n}$ is a bivariate Laurent polynomial defined as
\begin{align}
	H(z_m,z_n) = \sum_{k_m=k_{0,m}}^{k_{1,m}-1} \sum_{k_n=k_{0,n}}^{k_{1,n}-1} \, h_{k_m,k_n} \, z_m^{-k_m} z_n^{-k_n},
\end{align}
where $m$ refers to the horizontal axis and $n$ to the vertical one.
Moreover, to keep consistency with other papers, the $H^*(z_m,z_n) = H(z_n,z_m)$ denotes a filter transposed to the $H(z_m,z_n)$.
For simplicity, we have made a small abuse of notation.
Instead of the full notation $H(z_m,z_n)$, we only use a shortened labeling, such as $\mathrm{H}$.
Finally, we work with $2\times2$ and $4\times4$ matrices of Laurent polynomials.
These are usually referred to as the polyphase matrices.
The $2\times2$ matrices refers to the \mbox{1-D} systems, whereas the $4\times4$ to the \mbox{2-D} ones.
For simplicity, a shortened labeling is used for matrices as well.
The superscript $T$ denotes the vector or matrix transposition.

\subsection{Discrete Wavelet Transform}

The discrete wavelet transform has undergone a gradual development \cite{Mallat2009} in the last few decades.
First, S. Mallat \cite{Mallat1989} demonstrated the multi-scale wavelet decomposition computed with a pyramidal algorithm based on convolutions with quadrature mirror filters.
In detail, the discrete wavelet transform splits the input signal $x_k$ into two components L and H, each subsampled by a factor of 2.
Both of these components can be computed by the discrete convolution with two FIR filters $G_0(z)$ and $G_1(z)$ followed by the subsampling.
However, such computation is usually not the fastest one.
The transform can also be represented by the polyphase matrix \cite{Strang1997}.
Using this representation, the input signal is initially split into the L, H components.
No calculation is performed so far.
After such splitting, the DWT
\begin{align}
	\y = \M \, \x.
\end{align}
is described by the $2\times2$ matrix M mapping the initial components
\def\ww{2.8em}
\begin{align}
	\mathbf{x} &= \begin{bmatrix} \w{\L} & \w{\H} \end{bmatrix}^T
\end{align}
onto the resulting ones
\def\ww{2.8em}
\begin{align}
	\mathbf{y} &= \begin{bmatrix} \w{\L} & \w{\H} \end{bmatrix}^T.
\end{align}
The polyphase matrix is initially assembled as a polynomial matrix
\def\ww{2.8em}
\begin{align}
	\M = \begin{bmatrix}
		\w{\mathrm{G}_{1,o}} &    \mathrm{G}_{1,e}  \\
		   \mathrm{G}_{0,o}  & \w{\mathrm{G}_{0,e}} \\
	\end{bmatrix},
\end{align}
where subscript $e$ refers to the even coefficients, whereas $o$ refers to the odd coefficients.

\subsection{Lifting Scheme}

As a next step, W. Sweldens \cite{Sweldens1996,Daubechies1998} showed how any discrete wavelet transform can be decomposed into a sequence of simple filtering steps.
These steps are referred to as the lifting steps, and the scheme is known as the lifting scheme.
The lifting scheme reduces the number of arithmetic operations by up to 50\,\%.
The lifting steps occur in $K$ pairs.
The first step is referred to as the predict and the second one to as the update.
It may happen that the very first step of the lifting scheme is missing and the sequence of steps starts with the update step.
Usually, the very last step has a different form compared to all the others.
This one is then called the scaling step.
\def\ww{2.0em}
\begin{align}
	\M =
	\begin{bmatrix}
		\w{\zeta} & 0 \\
		0     & \w{1/\zeta}
	\end{bmatrix}
	\prod_{k=K-1}^{0}
	\begin{bmatrix}
		\w{1} & \U^{(k)} \\
		    0 & \w{1}
	\end{bmatrix}
	\begin{bmatrix}
		\w{1}    & 0 \\
		\P^{(k)} & \w{1}
	\end{bmatrix},
\end{align}
where $\zeta$ is a non-zero scaling factor, $\P^{(k)}$ is the $k$th predict convolution operator, and $\U^{(k)}$ is the $k$th update convolution operator.
In this paper, we focus on a single pair of lifting steps.
We thus omit the $(k)$ superscript.
We also omit the scaling step, as the application of this step is trivial.

In parallel environments \cite{Rauber2013}, the processing of a single or several adjacent signal samples is mapped to independent processing units, commonly referred to as the threads.
To avoid race conditions (the behavior where the output is dependent on the sequence or timing of other threads), the threads must use some type of synchronization method.
In this paper, we will consider the use of memory barriers.
When we return to the lifting scheme, these barriers are usually required before each of the individual lifting steps.
However, certain form of the steps guarantees correctness of the calculation even without using the memory barrier between them.
In this paper, the barriers are indicated by the $\big|$ symbol placed in between the steps.
For example, $ \mathrm{M}_2 \big| \mathrm{M}_1 $ denotes a sequence of two steps -- the initial $\mathrm{M}_1$ and the subsequent $\mathrm{M}_2$ -- separated by the barrier.

The schemes presented above can be extended into two dimensions.
The most widely used \mbox{2-D} extension is Mallat's \cite{Mallat1989} \mbox{2-D} decomposition.
The transform is defined as the tensor product of \mbox{1-D} transforms.
At each scale of such decomposition, we obtain a quadruple of wavelet coefficients (LL, HL, LH, HH).

\subsection{Convolution and Polyphase Schemes}

Similarly to the \mbox{1-D} case, the transform can be computed using the convolution scheme.
Considering this case, one needs to convolve the input signal with four \mbox{2-D} FIR filters.
This operation is followed by the subsampling in both dimensions.
However, in practical implementations, the subsamplings are built into the convolutions in order to save arithmetic operations.
This scheme will further be labeled as {Convolution}.
In this scheme, no barrier is required at all.

Moreover, the \mbox{2-D} transform can be described by the polyphase matrix as well.
Using the polyphase representation, the input signal is initially split into the four polyphase components.
No calculation is performed so far.
Further, the \mbox{2-D} DWT is described by the $4\times4$ matrix M mapping the input components
\def\ww{2.8em}
\begin{align}
	\x &= \begin{bmatrix} \w{\LL} & \w{\HL} & \w{\LH} & \w{\HH} \end{bmatrix}^T
\end{align}
onto the final ones
\def\ww{2.8em}
\begin{align}
	\y &= \begin{bmatrix} \w{\LL} & \w{\HL} & \w{\LH} & \w{\HH} \end{bmatrix}^T.
\end{align}
Similarly to the \mbox{1-D} case, this can be written as
\begin{align}
	\y = \N^{\vphantom(}_{\P,\U} \, \big| \, \x,
\end{align}
where $\P,\U$ are \mbox{1-D} predict and update convolution operators.
Please notice the included initial barrier.
This scheme will further be called as {Polyphase}.

To define the \mbox{2-D} polyphase matrices, the predict and update operators must first be migrated into two dimensions.
Coupled together with filter transposition defined above, the two-dimensional counterparts of the operators are defined like follows.
\begin{align}
	\begin{bmatrix}
		\P \\
		\U \\
		\P^* \\
		\U^*
	\end{bmatrix}
	=
	\begin{bmatrix}
		{P}(z_m,z_n) \\
		{U}(z_m,z_n) \\
		{P}^*(z_m,z_n) \\
		{U}^*(z_m,z_n)
	\end{bmatrix}
	=
	\begin{bmatrix}
		P(z_m) \\
		U(z_m) \\
		P(z_n) \\
		U(z_n)
	\end{bmatrix}
\end{align}
Roughly speaking, the $\P$ and $\U$ denote the filters oriented along the horizontal axes, whereas the $\P^*$ and $\U^*$ denote the filters oriented along the vertical one.

\subsection{Notation}

\begin{figure}[b]
	\centering
	\hspace*{\fill}%
	\subfigure[]{%
		\def\svgwidth{.333333333333\linewidth}%
		\scriptsize%
		\input{block-diagram-Sweldens-viewport.tex}%
		\label{fig:annex-block}%
	}%
	\hspace*{\fill}%
	\subfigure[]{%
		\def\svgwidth{.333333333333\linewidth}%
		\raisebox{20pt}{\input{core-2-3-small.tex}}%
		\label{fig:annex-dataflow3}%
	}%
	\hspace*{\fill}%
	\subfigure[]{%
		\raisebox{40pt}{\includegraphics{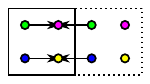}}%
		\label{fig:annex-dataflow2}%
	}\\
	\input{key.tex}
	\caption{
		Different visual representations of the same polyphase matrix.
	}
	\label{fig:annex}
\end{figure}

For readers not familiar with signal-processing notations, a relationship of the block and data-flow diagrams is explained in this section.
In this paper, we work with $4\times4$ matrices of Laurent polynomials, usually referred to as the polyphase matrices, for example, this one
\def\ww{2.8em}
\begin{align}
	\w{\T^H_{\P}} =
	\begin{bmatrix}
		\w{1} & 0     & 0     & 0     \\
		\P    & \w{1} & 0     & 0     \\
		0     & 0     & \w{1} & 0     \\
		0     & 0     & \P    & \w{1} \\
	\end{bmatrix}.
\end{align}
Since these matrices define a linear mapping from vectors of form $\begin{bmatrix} \w{\LL} & \w{\HL} & \w{\LH} & \w{\HH} \end{bmatrix}^T$ to vectors of the same form,
we can simply illustrate this mapping by the block diagram in \Figure{fig:annex-block}.

Moreover, the matrices are composed of elementary lifting operators like
\def\ww{2.8em}
\begin{align}
	\w{P({z})} = -1/2 ( 1 + z^{-1} ).
\end{align}
If we substitute such particular polynomials into the matrix, the mapping gets a specific shape, as illustrated by the dataflow diagram in \Figure{fig:annex-dataflow3}.
The solid arrows correspond to multiplication by $-1/2$ along with subsequent summation.
The dotted arrows similarly correspond to multiplication by factor of $1$, since the matrix $\T^H_{\P}$ contains ones on the main diagonal.

For reader's convenience, we use two-dimensional diagrams to illustrate the schemes with CDF 5/3 wavelets.
For the example above, such a diagram is shown in \Figure{fig:annex-dataflow2}, whereas the elementary quadruples of coefficients are highlighted by solid and dotted boxes.

\subsection{Sweldens Scheme}

\begin{figure}
	\centering
	{
		\def\svgwidth{\linewidth}
		\scriptsize
		\input{block-diagram-Sweldens.tex}
	}%
	\caption{
		Block diagram of the {Sweldens} scheme.
		The dashed vertical lines indicate barriers.
		The left half corresponds to the spatial predict operator, whereas the right half to the update one.
	}
	\label{fig:block-diagram-Sweldens}

	\vspace*{\floatsep}%

	\centering
	\subfigure[$\T^H_{\P}$]{\includegraphics{M_P_H}}
	\subfigure[$\T^V_{\P}$]{\includegraphics{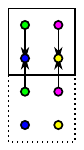}}
	\subfigure[$\S^H_{\U}$]{\includegraphics{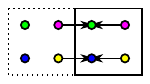}}
	\subfigure[$\S^V_{\U}$]{\includegraphics{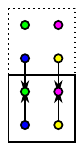}}
	\input{key.tex}
	\caption{
		\mbox{2-D} dataflow diagram, CDF 5/3 wavelet, {Sweldens} lifting scheme.
		The displayed part of the calculation results in the coefficients inside of the solid box.
		The dotted boxes refer to the surrounding threads.
	}
	\label{fig:dataflow-Sweldens}
\end{figure}

Following the Mallat's scheme, the predict and update lifting steps are applied in both directions sequentially.
This can be classified as a separable scheme.
As the convolution is the linear operator, horizontal and vertical steps can be arbitrary interleaved.
The baseline formulation of this scheme will be considered as follows.
The predict steps are always preceding the update ones.
Such separable scheme can be formally described by
\begin{align}
	\y = \S^V_{\U} \, \big| \, \S^H_{\U} \, \big| \, \T^V_{\P} \, \big| \, \T^H_{\P}  \, \big| \, \x,
\end{align}
where the individual matrices are defined as follows.
Let us mention a short comment on the matrix notation used.
For example, the matrix $\T^H_{\P}$ is parameterized by the $\P$ polynomial.
Further in the text, the same matrix appears parameterized by different polynomials, which is completely valid.
As it can be expected, the matrix $\T^H$ definition is not repeated for such case.
For better understanding, the corresponding signal-processing block diagram is shown in \Figure{fig:block-diagram-Sweldens}.
For the CDF 5/3 wavelet, these steps are also graphically illustrated in \Figure{fig:dataflow-Sweldens}.

\def\ww{2.8em}
\begin{align}
	\w{\T^H_{\P}} = 
	\begin{bmatrix}
		\w{1} & 0     & 0     & 0     \\
		\P    & \w{1} & 0     & 0     \\
		0     & 0     & \w{1} & 0     \\
		0     & 0     & \P    & \w{1} \\
	\end{bmatrix}
\end{align}

\def\ww{2.8em}
\begin{align}
	\w{\T^V_{\P}} =
	\begin{bmatrix}
		\w{1} & 0     & 0     & 0     \\
		0     & \w{1} & 0     & 0     \\
		\P^*  & 0     & \w{1} & 0     \\
		0     & \P^*  & 0     & \w{1} \\
	\end{bmatrix}
\end{align}

\def\ww{2.8em}
\begin{align}
	\w{\S^H_{\U}} =
	\begin{bmatrix}
		\w{1} & \U    & 0     & 0     \\
		0     & \w{1} & 0     & 0     \\
		0     & 0     & \w{1} & \U    \\
		0     & 0     & 0     & \w{1} \\
	\end{bmatrix}
\end{align}

\def\ww{2.8em}
\begin{align}
	\w{\S^V_{\U}} =
	\begin{bmatrix}
		\w{1} & 0     & \U^*  & 0     \\
		0     & \w{1} & 0     & \U^*  \\
		0     & 0     & \w{1} & 0     \\
		0     & 0     & 0     & \w{1} \\
	\end{bmatrix}
\end{align}

Please note the barriers in between each of the lifting steps.
In total, four barriers are required for each pair of the original \mbox{1-D} lifting steps.
This scheme will further be labeled as {Sweldens}.

Contemporary approaches on parallel architectures most commonly reflect this separable {Sweldens} scheme.
Exceptionally, the {Convolution} scheme is employed.
Considering the independent horizontal and vertical filtering steps, several different strategies of \mbox{2-D} DWT implementation can be used.
These strategies can be divided into three groups -- row--column, block-based, and pipelined methods.
The row--column methods process all of the horizontal filtering steps prior to the vertical ones.
The row--column method applied on the entire \mbox{2-D} image was used for instance in \cite{Franco2011,Tenllado2004,Tenllado2008,Franco2009,Blazewicz2012,Galiano2011,Galiano2013}.
In some papers, the transition between the horizontal and vertical stage is accompanied with data transposition.
The pipelined methods was used, e.g., in \cite{Laan2009} and \cite{Laan2011}.
These methods uses moving window for the vertical part of the transform.
However, the horizontal and vertical parts remain separated.
The block-based methods were used, e.g., in \cite{Matela2009,Blazewicz2012,Arguello2012,Song2014}.
The transform is tiled into blocks, in which the horizontal and vertical processing still remain separated.
However, between these parts, the data remain loaded in the local memory (making them faster accessible).
For the sake of completeness, some of the works compute an entire \cite{Matela2009} or partial \cite{Arguello2012} multi-scale transform inside the blocks.

\begin{figure}[b]
	\centering
	{
		\def\svgwidth{\linewidth}
		\scriptsize
		\input{block-diagram-Polyphase.tex}
	}%
	\caption{
		Block diagram of the {Polyphase} scheme.
		The dashed vertical lines indicate implicit barrier.
	}
	\label{fig:block-diagram-Polyphase}

	\vspace*{\floatsep}%

	\centering
	\subfigure[$\N^{\protect\vphantom(}_{\P,\U}$]{\includegraphics{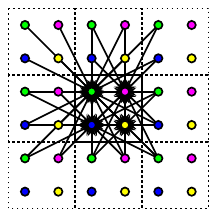}}
	\caption{
		\mbox{2-D} dataflow diagram, CDF 5/3 wavelet, {Polyphase} scheme.
		The solid box in the middle corresponds to the output coefficients.
	}
	\label{fig:dataflow-Polyphase}
\end{figure}

Going back to the {Polyphase} scheme, the polyphase matrix
\def\ww{2.8em}
\begin{align}
	\w{\N^{\vphantom(}_{\P,\U}} =
	\begin{bmatrix}
		\w{\V^*\V} & \V^*\U   & \U^*\V & \U^*\U \\
		\V^*\P     & \w{\V^*} & \U^*\P & \U^*   \\
		\P^*\V     & \P^*\U   & \w{\V} & \U     \\
		\P^*\P     & \P^*     & \P     & \w{1}  \\
	\end{bmatrix}
\end{align}
can expressed using the auxiliary polynomial $\V = \P\U + 1$.
The matrix can be obtained as the product of individual matrices of the {Sweldens} scheme.
In this scheme, it is no longer possible to distinguish the vertical and horizontal filtering.
Only an initial barrier is required for this scheme.
Unfortunately, the number of arithmetic operations has grown in proportion to the square of filter sizes.
The corresponding generic signal-processing diagram is shown in \Figure{fig:block-diagram-Polyphase}.
For the CDF 5/3 wavelet, these operations are illustrated in \Figure{fig:dataflow-Polyphase}.

\subsection{Iwahashi Scheme}

\begin{figure}
	\centering
	{
		\def\svgwidth{\linewidth}
		\scriptsize
		\input{block-diagram-Iwahashi.tex}
	}%
	\caption{
		Block diagram of the {Iwahashi} scheme.
		The dashed vertical lines indicate barriers.
	}
	\label{fig:block-diagram-Iwahashi}

	\vspace*{\floatsep}%

	\centering
	\subfigure[$\T^I_{\P}$]{\includegraphics{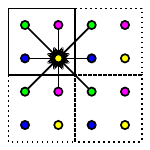}}%
	\subfigure[$\R^I_{\P,\U}$]{\includegraphics{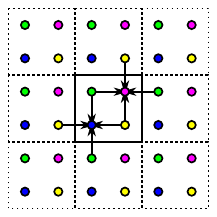}}%
	\subfigure[$\S^I_{\U}$]{\includegraphics{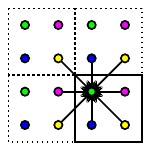}}%
	\caption{
		\mbox{2-D} dataflow diagram, CDF 5/3 wavelet, {Iwahashi} lifting scheme.
		The solid box corresponds to the output coefficients.
	}
	\label{fig:dataflow-Iwahashi}
\end{figure}

Recently, Iwahashi \textit{et al.} \cite{Iwahashi2007,Iwahashi2010,Iwahashi2013} presented the non\hyp{}separable lifting scheme, consisting of three spatial lifting steps.
As in the previous case, it is not possible to distinguish the vertical and horizontal filtering.
The three steps can be described as follows.
Initially, a \mbox{2-D} lifting step leading to the computation of the HH coefficient is performed.
This step corresponds to a spatial predict convolution operator.
This is followed by parallel computation of the HL and LH coefficients, using the original \mbox{1-D} predict and update filters.
In the third step, the LL coefficient is computed using another \mbox{2-D} filter.
The last step can be understood as a spatial update operator.
In the matrix notation, the scheme can be defined as
\begin{align}
	\y = \S^I_{\U} \, \big| \, \R^I_{\P,\U} \, \big| \, \T^I_{\P}  \, \big| \, \x,
\end{align}
where the individual matrices are defined as follows.
The signal-processing diagram is shown in \Figure{fig:block-diagram-Iwahashi}.
For the CDF 5/3 wavelet, the individual steps are illustrated in \Figure{fig:dataflow-Iwahashi}.

\def\ww{2.8em}
\begin{align}
	\w{\T^I_{\P}} =
	\begin{bmatrix}
		\w{1}  & 0     & 0     & 0     \\
		0      & \w{1} & 0     & 0     \\
		0      & 0     & \w{1} & 0     \\
		\P\P^* & \P^*  & \P    & \w{1} \\
	\end{bmatrix}
\end{align}

\def\ww{2.8em}
\begin{align}
	\w{\R^I_{\P,\U}} =
	\begin{bmatrix}
		\w{1} & 0     & 0     & 0     \\
		\P    & \w{1} & 0     & \U^*  \\
		\P^*  & 0     & \w{1} & \U    \\
		0     & 0     & 0     & \w{1} \\
	\end{bmatrix}
\end{align}

\def\ww{2.8em}
\begin{align}
	\w{\S^I_{\U}} =
	\begin{bmatrix}
		\w{1} & \U    & \U^*  & -\U\U^* \\
		0     & \w{1} & 0     & 0       \\
		0     & 0     & \w{1} & 0       \\
		0     & 0     & 0     & \w{1}   \\
	\end{bmatrix}
\end{align}

Three barriers are required in between these steps.
As for the {Polyphase} scheme, the number of arithmetic operations increased proportionally to the square of filter sizes.
However, the total number of operations is significantly lower.
This scheme will further be labeled as {Iwahashi}.

When we compare the separable {Sweldens} and non-separable {Iwahashi} schemes, some findings becomes obvious at first glance.
The number of operations tends to be considerably smaller for the separable case.
On the other hand, the number of memory barriers in the non-separable scheme was reduced to 75\,\% (from four to three barriers).
The {Polyphase} scheme stands apart from these two schemes.
It needs only an initial memory barrier.
Unfortunately, the number of arithmetic operations is unreasonably large.
This is caused by the number of non-zero elements in the corresponding polyphase matrix as well as by the degree of the longest filter $V$.
For clarification, the product of a Laurent polynomial of degree $|P(z)|$ and a Laurent polynomial of degree $|U(z)|$ is a Laurent polynomial of degree $|P(z)|+|U(z)|$.
Finally, the {Convolution} scheme employing four \mbox{2-D} filters is even worse in terms of the operations.
Anyway, only an initial memory is required here as well.
Detailed quantitative comparison is provided in \Section{sec:evaluation}.

When we consider the linearity of the convolution and the dependencies between the individual lifting steps, several gaps can be inferred in the schemes described above.
Recombining the operations into a new form could lead to the removal of unnecessary barriers.
Actually, exactly this idea is investigated in the following section, in which several novel \mbox{2-D} schemes are proposed.

Since this work is based on our previous work in \cite{Kula2016b}, it should be explained what the difference between this work and \cite{Kula2016b} is.
In \cite{Kula2016b}, we presented a block-based method employing a scheme foregoing the schemes proposed in this paper.
Unlike \cite{Kula2016b}, the schemes presented in this paper are defined by general predict and update operators.

\section{Proposed Schemes}
\label{sec:proposed-schemes}

In this section, the polyphase matrices, known so far, are reassembled in order to obtain the schemes suitable for parallel architectures.
All of the schemes discussed here are general, and they can be used for any discrete wavelet transform.
Please note that the contribution of this paper is presented in this section and the following one.

\subsection{Explosive Scheme}

\begin{figure}
	\centering
	{
		\def\svgwidth{\linewidth}
		\scriptsize
		\input{block-diagram-Explosive.tex}
	}%
	\caption{
		Block diagram of the {Explosive} scheme.
		The dashed vertical lines indicate barriers.
	}
	\label{fig:block-diagram-Explosive}

	\vspace*{\floatsep}%

	\centering
	\subfigure[$\T^E_{\P}$]{\includegraphics{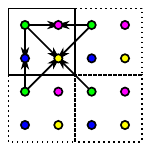}}%
	\subfigure[$\R^E_{\P,\U}$]{\includegraphics{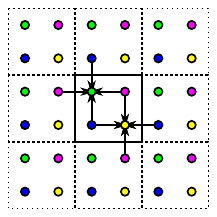}}%
	\subfigure[$\S^E_{\U}$]{\includegraphics{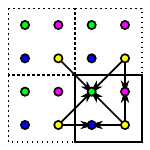}}%
	\caption{
		\mbox{2-D} dataflow diagram, CDF 5/3 wavelet, {Explosive} lifting scheme.
		The solid box corresponds to the output coefficients.
	}
	\label{fig:dataflow-Explosive}
\end{figure}

When we take a detailed look at the original \mbox{1-D} lifting scheme, a certain pattern can be identified in the predict and update steps.
Particularly, the predicts transmit data from L into H samples, whereas the updates transmit data from H into L.
The transmission can be viewed from two perspectives -- the data flow out from a source component (similarly to an explosion); or the data flow in into a destination component (an implosion).
As it can be expected, the {Sweldens} scheme exactly follows this pattern, since this scheme is a mere extension of \mbox{1-D} lifting into two dimensions.
Roles of source and destination samples properly turn during four lifting steps (horizontal and vertical, predict and update).
This procedure can be also seen as a data transmission in direction from LL into HH component (using \mbox{1-D} predicts), and a transmission from HH into LL one (using updates).
The HL and LH components are not relevant in this view.
The situation is clearly visible in \Figure{fig:block-diagram-Sweldens}.
In contrast to this scheme, the {Iwahashi} scheme has a different structure.
The leading step transmits data into HH component (using predicts), while the trailing one transmits them into LL one (updates).
However, no exclusive source components can be identified in this case.
The remaining step in the middle is not relevant.
See the block diagram in \Figure{fig:block-diagram-Iwahashi}.
Regarding to the perspectives outlined above, the {Iwahashi} scheme can be classified as an implosive one.
However, this is not the only three-step version (two-step scheme is discussed below in the text).
Similar scheme can be formulated using data explosions instead of the implosions.
Particularly, the LL component spreads the data into its neighborhood during the predict step, whereas the data flow out from the HH component in the update step.
No exclusive destination components can be identified here as well.
Again, the step in the middle is not relevant.
For further purposes, this newly proposed scheme will be labeled as {Explosive}.
The block diagram is shown in \Figure{fig:block-diagram-Explosive}.
The steps for the CDF 5/3 wavelet are also illustrated in \Figure{fig:dataflow-Explosive}.
Formally, the scheme can be defined as
\begin{align}
	\y = \S^E_{\U} \, \big| \, \R^E_{\P,\U} \, \big| \, \T^E_{\P}  \, \big| \, \x,
\end{align}
where the individual matrices follows.
Three barriers are required, as in the case of the {Iwahashi} scheme.

\def\ww{2.8em}
\begin{align}
	\w{\T^E_{\P}} =
	\begin{bmatrix}
		\w{1}  & 0      & 0      & 0      \\
		\P     & \w{1}  & 0      & 0      \\
		\P^*   & 0      & \w{1}  & 0      \\
		-\P\P^*& 0      & 0      & \w{1}  \\
	\end{bmatrix}
\end{align}

\def\ww{2.8em}
\begin{align}
	\w{\R^E_{\P,\U}} =
	\begin{bmatrix}
		\w{1}  & \U     & \U^*   & 0      \\
		0      & \w{1}  & 0      & 0      \\
		0      & 0      & \w{1}  & 0      \\
		0      & \P^*   & \P     & \w{1}  \\
	\end{bmatrix}
\end{align}

\def\ww{2.8em}
\begin{align}
	\w{\S^E_{\U}} =
	\begin{bmatrix}
		\w{1}  & 0      & 0      & \U\U^* \\
		0      & \w{1}  & 0      & \U^*   \\
		0      & 0      & \w{1}  & \U     \\
		0      & 0      & 0      & \w{1}  \\
	\end{bmatrix}
\end{align}

\subsection{Monolithic Scheme}

\begin{figure}[b]
	\centering
	{
		\def\svgwidth{\linewidth}
		\scriptsize
		\input{block-diagram.tex}
	}%
	\caption{
		Block diagram of the {Monolithic} scheme.
		The dashed vertical lines indicate barriers.
		The left half corresponds to the predict operator, whereas the right half to the update.
	}
	\label{fig:block-diagram-Proposed-1}

	\vspace*{\floatsep}%

	\centering
	\hspace*{\fill}%
	\subfigure[$\T^{\protect\vphantom(}_{\P}$]{\includegraphics{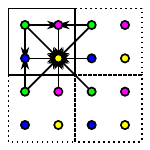}}%
	\hspace*{\fill}%
	\subfigure[$\S^{\protect\vphantom(}_{\U}$]{\includegraphics{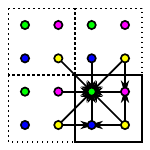}}%
	\hspace*{\fill}%
	\caption{
		\mbox{2-D} dataflow diagram, CDF 5/3 wavelet, {Monolithic} scheme.
		The solid box corresponds to the output coefficients.
	}
	\label{fig:dataflow-Proposed-1}
\end{figure}

Motivated by the work of Iwahashi \textit{et al.} \cite{Iwahashi2013}, we have reorganized the elementary lifting filters in order to remove the middle lifting step.
This action consequently reduces the number of memory barriers.
As a result, we receive a new two-step non-separable scheme.
The first step corresponds to a spatial predict operator.
This one is completely responsible for the HH coefficient.
In addition, the HL and LH coefficients are partially computed here as well.
The second step corresponds to a spatial update.
It is responsible for the LL coefficient and completion of the HL and LH ones.
Formally, the scheme is defined as
\begin{align}
	\y = \S^{\vphantom(}_{\U} \, \big| \, \T^{\vphantom(}_{\P}  \, \big| \, \x,
\end{align}
where the $\S_{\U}$ and $\T_{\P}$ are defined as follows.
Moreover, the hypotetical signal-processing diagram is shown in \Figure{fig:block-diagram-Proposed-1}.
For the CDF 5/3 wavelet, the scheme is graphically illustrated in \Figure{fig:dataflow-Proposed-1}.

\begin{table*}
	\def\Scale{1}%
	\newcolumntype{C}{ >{\centering\arraybackslash} m{2.2cm} }%
	\centering
	\begin{tabu} to \linewidth {C | C C | C | C | C }
		\toprule
			step\vphantom{\big(}
			& \multicolumn{2}{c|}{Sweldens}
			& {Monolithic}\vphantom{\big(}
			& {Iwahashi}\vphantom{\big(}
			& {Explosive}\vphantom{\big(} \\
		\midrule
			predict
			& \adjustbox{trim={.20\width} {.20\height} {0.20\width} {.20\height},clip}{\includegraphics[scale=\Scale]{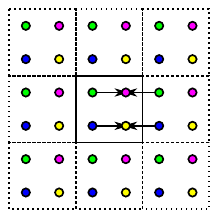}}
			& \adjustbox{trim={.20\width} {.20\height} {0.20\width} {.20\height},clip}{\includegraphics[scale=\Scale]{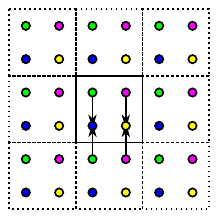}}
			& \adjustbox{trim={.20\width} {.20\height} {0.20\width} {.20\height},clip}{\includegraphics[scale=\Scale]{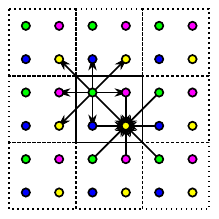}}
			& \adjustbox{trim={.20\width} {.20\height} {0.20\width} {.20\height},clip}{\includegraphics[scale=\Scale]{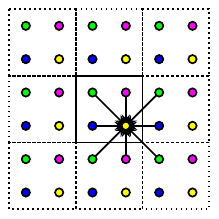}}
			& \adjustbox{trim={.20\width} {.20\height} {0.20\width} {.20\height},clip}{\includegraphics[scale=\Scale]{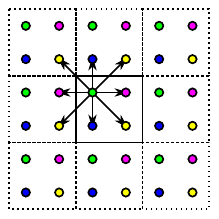}} \\
		\midrule
			{middle}
			& 
			& 
			& 
			& \adjustbox{trim={.20\width} {.20\height} {0.20\width} {.20\height},clip}{\includegraphics[scale=\Scale]{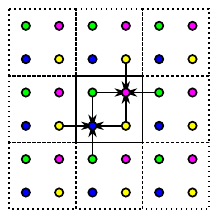}}
			& \adjustbox{trim={.20\width} {.20\height} {0.20\width} {.20\height},clip}{\includegraphics[scale=\Scale]{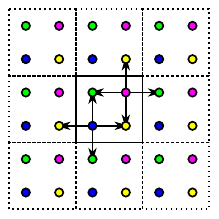}} \\
		\midrule
			update
			& \adjustbox{trim={.20\width} {.20\height} {0.20\width} {.20\height},clip}{\includegraphics[scale=\Scale]{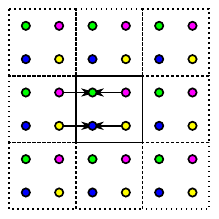}}
			& \adjustbox{trim={.20\width} {.20\height} {0.20\width} {.20\height},clip}{\includegraphics[scale=\Scale]{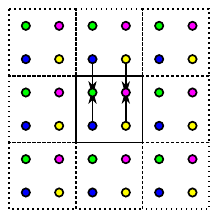}}
			& \adjustbox{trim={.20\width} {.20\height} {0.20\width} {.20\height},clip}{\includegraphics[scale=\Scale]{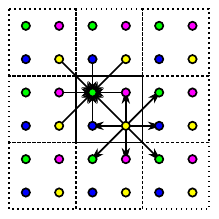}}
			& \adjustbox{trim={.20\width} {.20\height} {0.20\width} {.20\height},clip}{\includegraphics[scale=\Scale]{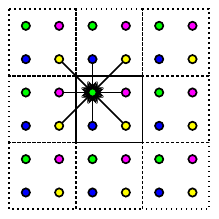}}
			& \adjustbox{trim={.20\width} {.20\height} {0.20\width} {.20\height},clip}{\includegraphics[scale=\Scale]{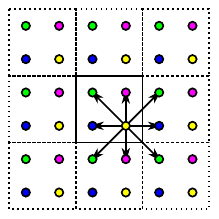}} \\
		\bottomrule
	\end{tabu}
	\caption{
		CDF 5/3 wavelet.
		Shapes of spatial lifting steps for selected schemes.
		The step in the middle raised from the combination of the original predict and update steps.
		Illustrative purpose only.
	}
	\label{tab:shapes}
\end{table*}

\def\ww{2.8em}
\begin{align}
	\w{\T^{\vphantom(}_{\P}} =
	\begin{bmatrix}
		\w{1}  & 0     & 0     & 0     \\
		\P     & \w{1} & 0     & 0     \\
		\P^*   & 0     & \w{1} & 0     \\
		\P\P^* & \P^*  & \P    & \w{1} \\
	\end{bmatrix}
\end{align}
\def\ww{2.8em}
\begin{align}
	\w{\S^{\vphantom(}_{\U}} =
	\begin{bmatrix}
		\w{1} & \U    & \U^*  & \U\U^* \\
		0     & \w{1} & 0     & \U^*   \\
		0     & 0     & \w{1} & \U     \\
		0     & 0     &     0 & \w{1}  \\
	\end{bmatrix}
\end{align}
The total number of operations remained the same as for the {Iwahashi} scheme.
However, the number of the explicit barriers has been reduced to only two.
This is a crucial contribution of our work.
Further in the paper, this scheme will be labeled as {Monolithic}.
One can easily verify the correctness of the proposed scheme by comparing the product $\S^{\vphantom(}_{\U} \T^{\vphantom(}_{\P}$ to the matrix $\N^{\vphantom(}_{\P,\U}$ of the {Polyphase} scheme.

A comparison of the shapes for selected schemes can be found in \Table{tab:shapes}.
Regarding the {Polyphase} scheme, no spatial predict nor update step can be identified in its calculation.

In practical implementations, the formed intermediate coefficients cannot take the same place as the input ones.
Otherwise, the race condition occurs.
This implies a higher memory consumption compared to the previous schemes.
A particular numbers are listed in \Table{tab:scheme-memory-barriers}.

Two simple observations can be made from the scheme presented so far.
The {Sweldens} scheme requires the lowest number of operations.
In contrast to this approach, the non-separable scheme proposed above requires the lowest number of memory barriers.
Combining these two observations together, new schemes can be formed.
This possibility is investigated below.

\section{Improvements}
\label{sec:improvements}

Additionally, we have made another observation.
The operation composed as a product of monomials with the exponent of $z_n$ and $z_n$ being equal to zero (i.e., scalars) never touch the coefficients belonging to the surrounding threads.
As the convolution is the linear operation, this monomial can be detached from the original operator and subsequently calculated using the {Sweldens} scheme.
This scheme has a minimal number of arithmetic operations.
The rest of the original polynomial shall be computed using different scheme, according to suitability for a particular platform.

\begin{figure}[b]
	\centering
	\subfigure[$\T^H_{\P_0}$]{\includegraphics{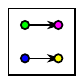}}
	\subfigure[$\T^V_{\P_0}$]{\includegraphics{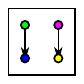}}
	\subfigure[$\S^H_{\U_0}$]{\includegraphics{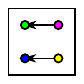}}
	\subfigure[$\S^V_{\U_0}$]{\includegraphics{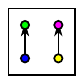}}
	\caption{
		\mbox{2-D} dataflow diagram, CDF 5/3 wavelet, common steps for all improved schemes.
	}
	\label{fig:dataflow-*-Improved}
\end{figure}

In more detail, the original filters were split into two halves as $\P = \P_0 + \P_1$, and $\U = \U_0 + \U_1$, where $\P_0$ and $\U_0$ are scalars.
This is a fundamental step for the following constructions.
Now, the scalars $\P_0,\U_0$ can be utilized in the separable {Sweldens} scheme.
This part will never touch the extraneous threads.
For a better understanding, see the dataflow diagram in \Figure{fig:dataflow-*-Improved}.
Conversely, the $\P_1,\U_1$ shall be employed in the {Explosive}, {Iwahashi}, {Monolithic}, or {Polyphase} scheme in order to minimize the number of required memory barriers.
Note that these two schemes can be combined into joint lifting steps.
However, such optimization is a simple matter of a specific implementation.

\begin{figure}
	\centering
	\subfigure[$\T^I_{\P_1}$]{\includegraphics{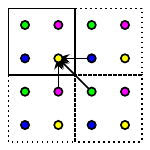}}%
	\subfigure[$\R^I_{\P_1,\U_1}$]{\includegraphics{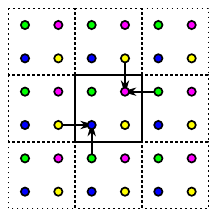}}%
	\subfigure[$\S^I_{\U_1}$]{\includegraphics{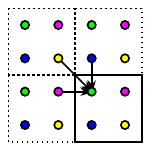}}%
	\caption{
		\mbox{2-D} dataflow diagram, CDF 5/3 wavelet, {Iwahashi$^*$} scheme.
		The solid box corresponds to the output coefficients.
	}
	\label{fig:dataflow-Iwahashi-Improved}
\end{figure}

Initially, we have employed the idea described in previous paragraphs in conjunction with the {Iwahashi} scheme.
The resulting scheme is defined as
\begin{align}
	\y = \S^V_{\U_0} \, \S^H_{\U_0} \, \S^I_{\U_1} \, \big| \, \R^I_{\P_1,\U_1} \, \big| \, \T^I_{\P_1}  \, \big| \, \T^V_{\P_0} \, \T^H_{\P_0} \x,
\end{align}
where the individual matrices are defined above in the paper.
The number of barriers remains the same as for the original {Iwahashi} scheme.
The operations represented by the matrices defined for the {Sweldens} scheme do not need to be preceded by a barrier.
The scheme will be further referred to as {Iwahashi$^*$}.
For the CDF 5/3 wavelet, this scheme is graphically illustrated in \Figure{fig:dataflow-Iwahashi-Improved}.

\begin{figure}
	\centering
	\subfigure[$\T^E_{\P_1}$]{\includegraphics{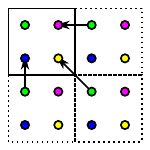}}%
	\subfigure[$\R^E_{\P_1,\U_1}$]{\includegraphics{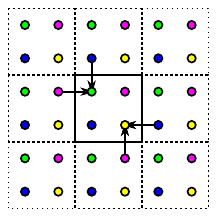}}%
	\subfigure[$\S^E_{\U_1}$]{\includegraphics{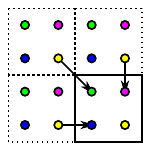}}%
	\caption{
		\mbox{2-D} dataflow diagram, CDF 5/3 wavelet, {Explosive$^*$} scheme.
		The solid box corresponds to the output coefficients.
	}
	\label{fig:dataflow-Explosive-Improved}
\end{figure}

Similarly, we have employed the same trick in conjunction with the {Explosive} scheme.
This time, the scheme is defined as
\begin{align}
	\y = \S^V_{\U_0} \, \S^H_{\U_0} \S^E_{\U_1} \, \big| \, \R^E_{\P_1,\U_1} \, \big| \, \T^E_{\P_1} \, \big| \, \T^V_{\P_0} \, \T^H_{\P_0} \, \x.
\end{align}
Also in this case, the number of barriers remains the same as for the original scheme.
Analogously to the previous case, this scheme will be referred to as {Explosive$^*$}.
The dataflow diagram for the CDF 5/3 wavelet is shown in \Figure{fig:dataflow-Explosive-Improved}.

\begin{figure}
	\centering
	\hspace*{\fill}%
	\subfigure[$\T^{\protect\vphantom(}_{\P_1}$]{\includegraphics{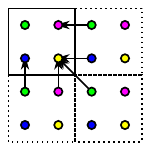}}%
	\hspace*{\fill}%
	\subfigure[$\S^{\protect\vphantom(}_{\U_1}$]{\includegraphics{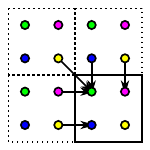}}%
	\hspace*{\fill}%
	\caption{
		\mbox{2-D} dataflow diagram, CDF 5/3 wavelet, {Monolithic$^*$} scheme.
		The solid box corresponds to the output coefficients.
	}
	\label{fig:dataflow-Proposed-2}
\end{figure}

As a next step, consider a new construction based on the {Monolithic} scheme.
The same trick can be utilized here as well.
In the matrix notation, the newly composed scheme is defined as
\begin{align}
	\y = \S^V_{\U_0} \, \S^H_{\U_0} \, \S^{\vphantom(}_{\U_1} \, \big| \, \T^V_{\P_0} \, \T^H_{\P_0} \, \T^{\vphantom(}_{\P_1}  \, \x,
\end{align}
where the individual matrices are defined above in the text.
For the CDF 5/3 wavelet, this scheme is graphically illustrated in \Figure{fig:dataflow-Proposed-2}.
We will label this scheme as {Monolithic$^*$}.

The schemes described above are formed such a way that the first lifting step (comprising $\P_1, \U_1$) after the barrier access coefficients of the surrounding threads.
The subsequent or preceding steps (comprising $\P_0, \U_0$) read only the local coefficients, which are not accessed by the other threads.
Then, the whole sequence can be repeated.
Of course, the calculation of transforms consisting of several pairs of lifting steps comprises several such connected schemes.

\begin{figure}
	\centering
	\subfigure[$\N^{\protect\vphantom(}_{\P_1,\U_1}$]{\includegraphics{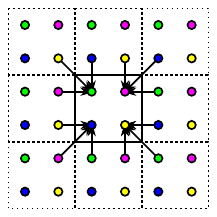}}
	\caption{
		\mbox{2-D} dataflow diagram, CDF 5/3 wavelet, \mbox{Polyphase$^*$} scheme.
		The solid box corresponds to the output.
	}
	\label{fig:dataflow-Proposed-3}
\end{figure}

Finally, we have decided to remove the last explicit barrier, leaving only the initial one in place.
The trick lies in the appropriate combination of the {Sweldens} and {Polyphase} schemes.
This time, the non-separable parts are merged into a joint step $\N_{\P_1,\U_1}$.
This step is inherently preceded by a barrier.
In case of an initial pair of lifting steps, the barrier at the beginning of the computation is used for this purpose.
In more detail, after the input data have been read by each computation unit, the calculations $\T^V_{\P_0} \, \T^H_{\P_0}$ are immediately performed.
At this point, the intermediate results can be appropriately shared.
This is followed by the initial barrier.
Regarding the transforms consisting of several such schemes, the barrier between the connecting schemes is gratefully exploited.
In any case, the scheme is thus composed as
\begin{align}
	\y = \S^V_{\U_0} \, \S^H_{\U_0} \, \N^{\vphantom(}_{\P_1,\U_1} \, \big| \, \T^V_{\P_0} \, \T^H_{\P_0}  \, \x
\end{align}
including the discussed barrier.
For the CDF 5/3 wavelet, the steps are illustrated in \Figure{fig:dataflow-Proposed-3}.
We will label this scheme as {Polyphase$^*$}.

For the sake of clarity, the proposed schemes will now be summarized.
By reversing the direction of filtering steps in the {Iwahashi} scheme, the new {Explosive} scheme was formed.
As a next step, the polynomials of the original polyphase matrix were reassembled into a new two-step form.
In between the steps, a memory barrier has to be placed.
This scheme is denoted as {Monolithic}.
Moreover, the number of arithmetic operations was reduced by splitting the polynomial into two parts.
These newly formed polynomials are then employed in appropriate schemes.
In this manner, the number of barriers remains unaffected, while the number of operations has been reduced.
This simple trick has resulted in the {Iwahashi$^*$}, {Explosive$^*$}, {Monolithic$^*$}, and {Polyphase$^*$} schemes.
Once again, we would like to emphasize that the schemes presented in this paper are general and they are not limited to any specific type of transform.

\section{Evaluation}
\label{sec:evaluation}

This section analyzes in detail various attributes of the schemes described in the previous sections.
Namely, synchronization and memory demands for different wavelets are examined.
We realize that such properties do not provide sufficient information on a performance in real environments.
For this reason, we are interested in comparing the performance of the discussed schemes on real graphics cards in terms of memory bandwidth in the next section.

The evaluation is presented using the following three wavelets.
The first wavelet we have employed is the CDF \cite{Cohen1992} 5/3 wavelet.
This one is used for a lossless compression in the JPEG 2000 compression standard.
The lifting scheme is defined by
\begin{align}
	\begin{bmatrix}
		P({z}) \\
		U({z})
	\end{bmatrix}
	=
	\begin{bmatrix}
		-1/2 ( 1 + z^{-1} ) \\
		\phantom{+}1/4 ( 1 + z^{\phantom{+1}})
	\end{bmatrix},
\end{align}
and the scaling factor $ \zeta = \sqrt{2}. $

As the second wavelet, we have chosen the CDF 9/7 wavelet.
In the JPEG 2000 standard, this wavelet is used as a basis for a lossy compression.
The underlying scheme is given by
\begin{align}
	\begin{bmatrix}
		{P}^{(0)}({z}) \\
		{U}^{(0)}({z}) \\
		{P}^{(1)}({z}) \\
		{U}^{(1)}({z})
	\end{bmatrix}
	=
	\begin{bmatrix}
		\alpha ( 1 + z^{-1} ) \\
		\beta  ( 1 + z^{\phantom{+1}}) \\
		\gamma ( 1 + z^{-1} ) \\
		\delta ( 1 + z^{\phantom{+1}}) \\
	\end{bmatrix},
\end{align}
where the $ \alpha, \beta, \gamma, \delta, $ and the $ \zeta $ are defined in \cite{Daubechies1998}.
Both the CDF wavelets have predict and update convolution operators of degree 1 (two-tap symmetric filters).

The last wavelet included in the comparison is $(4,4)$ interpolating transform built from the interpolating Des\-lau\-ri\-ers--Dubuc \cite{Sweldens1996}, defined by
\begin{align}
	\begin{bmatrix}
		P({z}) \\
		U({z})
	\end{bmatrix}
	=
	\begin{bmatrix}
		1/16( z + z^{-2} ) -9/16( 1      + z^{-1} ) \\
		9/32( 1 + z      ) -1/32( z^{-1} + z^{2}  )
	\end{bmatrix}.
\end{align}
This wavelet is used in Dirac video compression standard.
For simplicity, we refer this one to as DD 13/7.
The underlying lifting scheme differs from the two previous in employed predict and update convolution operators.
These operators now have a degree of 3 instead of 1.
Consequently, this difference has resulted in a significantly higher number of arithmetic operations in the case of non-separable filtering steps.

\begin{table}
	\centering
	\begin{tabu} to \linewidth {X[l]X[l]|X[r]X[r]}
		\toprule
		wavelet
		& scheme                & barriers & operations \\
		\midrule
		\multirow{10}{*}{CDF 5/3}
		& {Sweldens}            &        4 &      16 \\ 
		& {Iwahashi}            &        3 &      24 \\ 
		& {Iwahashi$^*$     }   &        3 &      18 \\ 
		& {Explosive}           &        3 &      24 \\ 
		& {Explosive$^*$     }  &        3 &      18 \\ 
		& {Monolithic}          &        2 &      24 \\ 
		& {Monolithic$^*$     } &        2 &      18 \\ 
		& {Polyphase}           &        1 &      63 \\ 
		& {Polyphase$^*$     }  &        1 &      23 \\ 
		& {Convolution}         &        1 &      64 \\ 
		\midrule
		\multirow{10}{*}{CDF 9/7}
		& {Sweldens}            &        8 &      32 \\ 
		& {Iwahashi}            &        6 &      48 \\ 
		& {Iwahashi$^*$     }   &        6 &      36 \\ 
		& {Explosive}           &        6 &      48 \\ 
		& {Explosive$^*$     }  &        6 &      36 \\ 
		& {Monolithic}          &        4 &      48 \\ 
		& {Monolithic$^*$     } &        4 &      36 \\ 
		& {Polyphase}           &        2 &     126 \\ 
		& {Polyphase$^*$     }  &        2 &      46 \\ 
		& {Convolution}         &        1 &     256 \\ 
		\midrule
		\multirow{10}{*}{DD 13/7}
		& {Sweldens}            &        4 &      32 \\ 
		& {Iwahashi}            &        3 &      64 \\ 
		& {Iwahashi$^*$     }   &        3 &      50 \\ 
		& {Explosive}           &        3 &      64 \\ 
		& {Explosive$^*$     }  &        3 &      50 \\ 
		& {Monolithic}          &        2 &      64 \\ 
		& {Monolithic$^*$     } &        2 &      50 \\ 
		& {Polyphase}           &        1 &     255 \\ 
		& {Polyphase$^*$     }  &        1 &     203 \\ 
		& {Convolution}         &        1 &     256 \\ 
		\bottomrule
	\end{tabu}
	\caption{
		Number of operations and memory barriers examined for various wavelets.
	}
	\label{tab:number-mac-barrier}
\end{table}

The first examined parameters include the number of arithmetic operations (the scaling steps were omitted) and the number of memory barriers.
The schemes presented in this paper can be directly applied on the CDF 5/3 and DD 13/7 transforms, as these comprises only a single pair of lifting steps.
The CDF 9/7 transform is computed by two such connected schemes.
The comparison is shown in \Table{tab:number-mac-barrier}.
Several expectations can be made from the table.
On architectures based on serial computation, the schemes should perform accordingly to the number of arithmetic operations.
However, on the parallel architectures, the number of employed memory barriers is expected to play an important role.
Some of the schemes could benefit from this property.

\begin{table}[b]
	\centering
	\begin{tabu} to \linewidth {l|X[r]X[r]X[r]}
		\toprule
		scheme                &   barriers &     single &     double \\
		\midrule
		{Sweldens}            &         4  & \textbf{2} & \textbf{3} \\
		{Iwahashi}            &         3  &         3  &         4  \\
		{Iwahashi$^*$     }   &         3  &         3  &     (6) 4  \\
		{Explosive}           &         3  & \textbf{2} & \textbf{3} \\
		{Explosive$^*$     }  &         3  & \textbf{2} & \textbf{3} \\
		{Monolithic}          &         2  &         3  &         6  \\
		{Monolithic$^*$     } &         2  &         3  &         6  \\
		{Polyphase}           & \textbf{1} &         4  &     (8) 4  \\
		{Polyphase$^*$     }  & \textbf{1} &         4  &     (8) 4  \\
		\bottomrule
	\end{tabu}
	\caption{
		Number of memory barriers and local memory cells per quadruple required by the schemes discussed in this paper.
		Memory cells are given for a single buffering (two barriers) as well as a double buffering (only a single barrier).
		The numbers in parentheses are valid in the case of connecting schemes.
		Best features in bold.
	}
	\label{tab:scheme-memory-barriers}
\end{table}

\begin{table}[b]
	\centering
	\begin{tabu} to \linewidth {l|X[r]X[r]X[r]}
		\toprule
		scheme                & write & read degree-1 & read degree-3 \\
		\midrule
		{Sweldens}            &$ 1+4K$&$           8K$&$      24K$\\
		{Iwahashi}            &$ 2+4K$&$          10K$&$      42K$\\
		{Iwahashi$^*$     }   &$   6K$&$          10K$&$      42K$\\
		{Explosive}           &$   4K$&$          10K$&$      42K$\\
		{Explosive$^*$     }  &$   4K$&$          10K$&$      42K$\\
		{Monolithic}          &$   6K$&$          10K$&$      42K$\\
		{Monolithic$^*$     } &$   6K$&$          10K$&$      42K$\\
		{Polyphase}           &$   4K$&$          21K$&$     117K$\\ 
		{Polyphase$^*$     }  &$   4K$&$          12K$&$     117K$\\
		\bottomrule
	\end{tabu}
	\caption{
		Number of local memory reads and writes for all schemes and wavelets under examination.
		The $K$ denotes the number of predict/update pairs.
		The degree-1 polynomials correspond to factorizations of CDF wavelets, whereas \mbox{degree-3} to DD 13/7.
	}
	\label{tab:scheme-memory}
\end{table}

As can be seen from the referenced table, the {Sweldens} scheme always leads to the smallest number of operations coupled with the highest number of barriers.
The recently proposed {Iwahashi} scheme reduces the number of barriers by one per one pair of original \mbox{1-D} lifting steps.
Unfortunately, the number of operations is increased at the same time.
This increase is particularly noticeable on longer lifting filters, as in the case of DD 13/7 wavelet.
The {Monolithic} scheme further reduces the number of barriers by one per one pair of original steps while keeping the number of operations untouched.
In addition to this, the {Monolithic$^*$} scheme reduces the number of operations.
This reduction is most evident on short lifting filters.
For instance, in the case of CDF wavelets, the number of operations is reduced to 75\,\%, whereas in the case of DD 13/7 wavelet, the number of operations is only reduced to 78\,\%.
The number of barriers per one pair of original lifting steps can be even further reduced to a single one by combining all operations into a single step.
Such case corresponds to the {Polyphase} scheme.
Unfortunately, the number of operations was increased enormously.
For shorter lifting filters, this number can be noticeably reduced using the {Polyphase$^*$} scheme, in which the number of barriers remains the same.
Finally, for lifting factorizations consisting of several pairs of steps, it makes sense to reduce the number of barriers to a single one by using the {Convolution} scheme.
In such case, the number of operations is sadly the highest of all of the schemes.

Other examined parameters included the memory footprint, and number of memory loads/stores.
These parameters can be determined from \Table{tab:scheme-memory-barriers} and \Table{tab:scheme-memory}.
All of the numbers are given with respect to the quadruple of coefficients, which usually correspond to a single thread.
The number of load (read) operations depends on the length of the lifting operators.
For example, the CDF 5/3 and CDF 9/7 factorizations consist of degree-1 convolutional filters.
On the contrary, the DD 13/7 consists of degree-3 filters.
The number of store (write) operations is independent of the underlying scheme.
It may happen that the local memory footprint for the connecting schemes ($K>1$) differs from the footprint for a single predict/update pair ($K=1$).
These numbers are indicated in the parentheses in \Table{tab:scheme-memory-barriers}.
For clarity, the number of memory barriers is not affected by the improvement proposed in \Section{sec:improvements}.

\section{Performance}
\label{sec:performance}

To evaluate the considered schemes, we have decided to use high-performance GPUs programmed using the OpenCL framework.
In terms of the OpenCL, the schemes are computed using parallel tasks referred to as the kernels.
One item from a collection of parallel executions of a kernel is referred to as the work-item or thread.
The threads that execute on a single compute unit are grouped into so-called work-groups.
The threads in the group execute the same kernel and share local memory.
Each work-group can synchronize the threads via memory barriers.
Work-groups cannot synchronize with each other.
Considering the processing of images, we map overlapping (in order to properly compute the coefficients near tile boundaries) image tiles onto the work-groups.
Moreover, each thread is responsible for a single quadruple of transform coefficients (LL, HL, LH, and HH).
At the beginning of the computation, the input image is placed in the global memory.
The tiles are then transferred into the local memory.
After the scheme computation, the resulting coefficients are copied back into the global memory.
Such strategy fulfills the definition of a single-loop data processing (therefore without unnecessary data transfers).

\begin{figure*}
	\subfigure[degree-1 schemes]{\includegraphics[width=.5\linewidth]{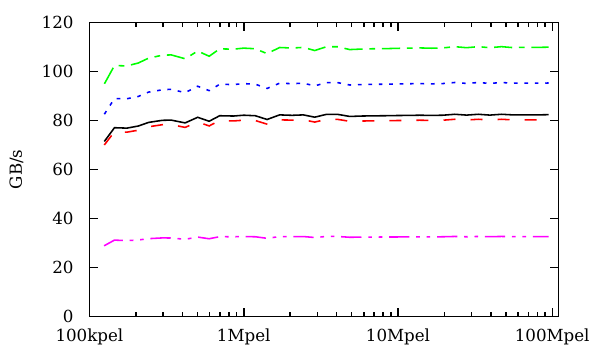}}
	\subfigure[degree-3 schemes]{\includegraphics[width=.5\linewidth]{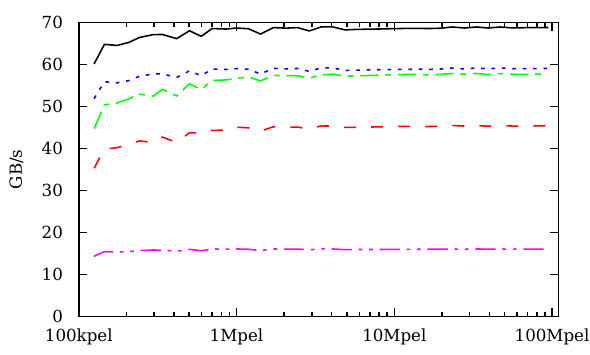}}\\
	\def\svgwidth{\linewidth}%
	\scriptsize
	\input{plot-none10-baseline-6970-key-2.tex}
	\caption{
		The baseline schemes on AMD 6970.
		Evaluation with the degree-1 and degree-3 lifting schemes.
		Only the performance of a transform code without the memory throughput was measured.
	}
	\label{fig:evaluation-none10-baseline-6970}
\end{figure*}

One needs to recall that the row--column, block-based, and pipelined methods denote an order in which an entire input image is processed.
The row--column approach indicates that all image rows are transformed prior to a transformation of all image columns (or vice versa).
Between these two parts, intermediate results are stored in the global memory.
However, all of the schemes presented in our paper are implemented as the block-based approaches.
This means that the entire input image is split into blocks (the tiles), which are then transformed at once using the local memory for storing the intermediate results.
Inside the blocks, some sort of separable scheme can be employed, which essentially corresponds to the row--column approach on a different scale.
The block-based approaches in various forms were also used in, e.g., \cite{Blazewicz2012,Arguello2012,Song2014,Kula2016b}.
Since the block-based approaches overcome the row--column ones (as shown in \cite{Kula2016b}, or analyzed in \cite{Song2014}), we do not include the classical row--column methods in our performance comparison.
Instead, we only compared different schemes employed under the block-based approach.
In this context, we would like to make a comment on data transfers between a device and host.
Due to the fact that the data are transferred in the same way for all schemes, we measured only a throughput based on a timing of a OpenCL kernels which calculate transforms.
Therefore, the transfer times between device and host are not our concern.

\begin{table}[b]
	\centering
	\begin{tabu} to \linewidth {l|X[r]X[r]}
		\multicolumn{1}{c}{~}
		                 &       AMD 6970 &       AMD 5870 \\
		\toprule
		vendor           &            AMD &            AMD \\
		model            & Radeon HD 6970 & Radeon HD 5870 \\
		\midrule
		VLIW length      &              4 &              5 \\
		multiprocessors  &             24 &             20 \\
		VLIW processors  &            384 &            320 \\
		total processors &         1\,536 &         1\,600 \\
		processor clock  &       880\,MHz &       850\,MHz \\
		performance      & 2\,703\,GFLOPS & 2\,720\,GFLOPS \\
		\midrule
		memory           &   1\,GiB GDDR5 &   1\,GiB GDDR5 \\
		memory clock     &    1\,375\,MHz &    1\,200\,MHz \\
		bandwidth        &      176\,GB/s &      154\,GB/s \\
		bus width        & \mbox{256-bit} & \mbox{256-bit} \\
		local memory     &        32\,KiB &        32\,KiB \\
		\bottomrule
	\end{tabu}
	\caption{
		Description of the GPUs used for the evaluation.
	}
	\label{tab:gpus}
\end{table}

The evaluation was performed primarily on two high-end GPUs --
AMD Radeon HD 6970 and AMD Radeon HD 5870.
Their technical parameters are summarized in \Table{tab:gpus}.
On both of the cards, variable length VLIW instructions are executed using blocks of 64 threads.
In more detail, VLIW instructions can be categorized into several groups (load/store instructions, barrier instructions, control flow instructions and ALU instructions).
To utilize whole processing capability, the VLIW instructions should be of maximal length.
In other words, as much as possible blocks of independent instructions should be presented in a kernel.

Several possibilities raised during the implementations of the presented schemes.
All of the schemes require several memory cells to interchange the intermediate coefficients.
Considering the GPUs, these coefficients can be efficiently stored in the local memory.
Unfortunately, it is not possible to rewrite these coefficients using a single memory barrier.
As a consequence, two possibilities occur -- double buffering using a single memory barrier, and single buffering using two of them.
The double buffering increases the memory requirements while maintaining the number of synchronizations.
Conversely, the single buffering introduces an addition barrier -- separating reading and rewriting of the coefficients.
For details, see \Table{tab:scheme-memory-barriers}.
In other words, one can choose whether intermediate results are overwritten in their place using two memory barriers or whether these are written to another location by making use of a single barrier.
Moreover, another possibility lies in the method of input and output data delivery.
For evaluation purposes, it is possible to completely omit the input and output of data.
The transform is not limited by memory bandwidth in this case.
For real scenarios, the data can be delivered using the global or texture memory.
In our experiments, we chose the latter option.

In the following paragraphs, three fundamental experiments on the described GPUs are presented.
The first experiment studies the performance of the baseline schemes mentioned in this paper.
The second experiment examines the influence of the improvement proposed in \Section{sec:improvements}.
Finally, the third experiment measures the real performance with CDF 9/7 wavelet and texture memory.

\begin{figure*}
	\subfigure[AMD 6970]{\includegraphics[width=.5\linewidth]{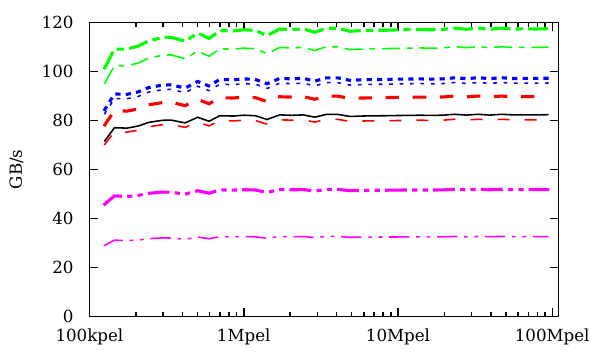}}%
	\subfigure[AMD 5870]{\includegraphics[width=.5\linewidth]{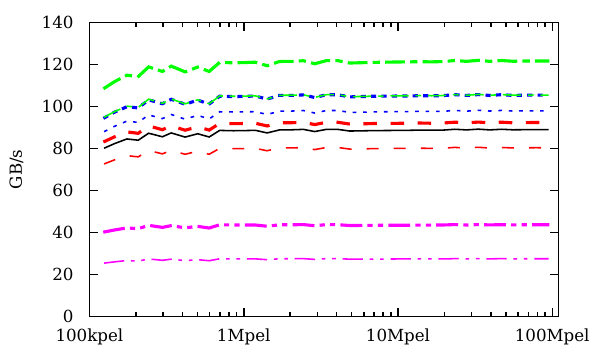}}\\
	\def\svgwidth{\linewidth}%
	\scriptsize
	\input{plot-none10-baseline+improved-6970+5870-key.tex}
	\caption{
		The schemes on AMD 6970 and AMD 5870.
		Evaluation with the degree-1 schemes.
		Only the performance of a transform code without the memory throughput was measured.
	}
	\label{fig:evaluation-none10-baseline+improved-6970+5870}
\end{figure*}

In the first experiment, the performance of the baseline schemes (without improvements proposed in \Section{sec:improvements}) was examined.
The measurements were conducted on the AMD 6970 card with two different lifting scheme shapes (degree-1 and degree-3 operators).
Only the transform performance was measured, without the influence of memory throughput.
The presented results are the average of ten measurements.
The results are shown in \Figure{fig:evaluation-none10-baseline-6970}.
One can easily observe a different behavior for short and long lifting operators.
For the short operators, the reduction in the number of lifting steps clearly improves the performance.
The situation actually corresponds directly to the number of memory barriers.
Conversely, in the case of the long operators, the situation is tilted in favor of the number of arithmetic operations.
Note that the horizontal axes are in a logarithmic scale.
The vertical axes express the transform throughput in GB/s (gigabytes per second).

\begin{figure*}
	\subfigure[AMD 6970]{\includegraphics[width=.5\linewidth]{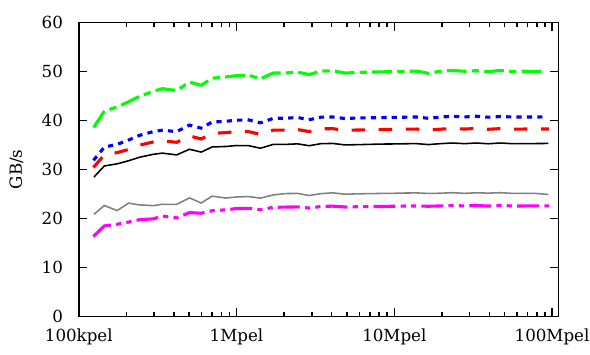}}%
	\subfigure[AMD 5870]{\includegraphics[width=.5\linewidth]{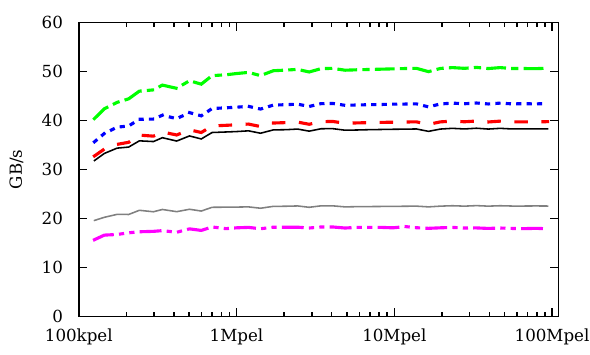}}\\
	\def\svgwidth{\linewidth}%
	\scriptsize
	\input{plot-texture-improved-6970+5870-key.tex}
	\caption{
		The improved schemes on AMD 6970 and AMD 5870.
		Evaluation with the CDF 9/7 wavelet and texture memory was performed.
	}
	\label{fig:evaluation-texture-improved-6970+5870}
\end{figure*}

In the second experiment, the contribution of the improvements proposed in \Section{sec:improvements} was examined.
The measurements were performed on both of the cards under the evaluation.
This time we have focused on the degree-1 schemes only.
As in the previous case, only the transform performance was measured using the average of ten measurements.
The results are shown in \Figure{fig:evaluation-none10-baseline+improved-6970+5870}.
As expected, the improvements slightly increase the transform performance.
However, the order of the schemes still corresponds to the number of memory barriers.
Several schemes perform even worse than the original separable {Sweldens} scheme -- namely, the original {Iwahashi} and both {Polyphase} schemes.
It is not surprising for the original {Polyphase} scheme, as this one exhibits quite a high number of operations and load instructions (see \Table{tab:scheme-memory} and \Table{tab:number-mac-barrier}).
In case of {Polyphase$^*$} scheme, the decisive factor was the number of load instructions coupled with a high local memory footprint (see \Table{tab:scheme-memory-barriers}).
A little surprising is the situation regarding the original {Iwahashi} scheme.
In this case, the scheme contains a relatively high number of operations, wherein there is no additional advantage.
For convenience, the values at the end of plots in \Figure{fig:evaluation-none10-baseline+improved-6970+5870} are listed in \Table{tab:evaluation-none10-baseline+improved-6970+5870}.

\begin{table}[!]
	\centering
	\begin{tabu} to \linewidth {l|X[r]X[r]}
		\toprule
			scheme           & throughput AMD~6970 & throughput AMD~5870 \\
		\midrule
			{Monolithic$^*$} &             117.426 &             121.579 \\
			{Monolithic}     &             109.865 &             105.407 \\
			{Explosive$^*$}  &              97.214 &             105.344 \\
			{Explosive}      &              95.263 &              97.877 \\
			{Iwahashi$^*$}   &              89.748 &              92.288 \\
			{Sweldens}       &              82.336 &              88.924 \\
			{Iwahashi}       &              80.284 &              80.283 \\
			{Polyphase$^*$}  &              51.776 &              43.619 \\
			{Polyphase}      &              32.593 &              27.462 \\
		\bottomrule
	\end{tabu}
	\caption{
		The degree-1 schemes on AMD 6970 and AMD 5870.
		The performance of a transform code without the memory throughput is listed.
		Values given in GB/s at the end of plots in \Figure{fig:evaluation-none10-baseline+improved-6970+5870}.
	}
	\label{tab:evaluation-none10-baseline+improved-6970+5870}
\end{table}

\begin{table*}
	\centering
	\begin{tabu} to \linewidth {l|X[r]X[r]X[r]|X[r]X[r]X[r]}
			\multicolumn{1}{c}{~} & \multicolumn{3}{c}{AMD 5870} & \multicolumn{3}{c}{AMD 6970} \\
		\toprule
			scheme           &  CDF 5/3 &  CDF 9/7 &  DD 13/7 &  CDF 5/3 &  CDF 9/7 &  DD 13/7 \\
		\midrule
			{Sweldens}       &    32.59 &    32.69 &    40.00 &    40.53 &    40.25 &    48.50 \\
			{Iwahashi}       &    34.88 &    35.90 &    50.12 &    41.33 &    42.33 &    61.80 \\
			{Iwahashi$^*$}   &    32.00 &    34.29 &    45.83 &    42.11 &    40.75 &    57.00 \\
			{Explosive}      &    36.88 &    39.32 &    49.57 &    45.14 &    46.64 &    60.47 \\
			{Explosive$^*$}  &    33.79 &    33.33 &    55.56 &    42.42 &    42.21 &    65.42 \\
			{Monolithic}     &    38.18 &    39.67 &    51.59 &    48.57 &    47.76 &    64.44 \\
			{Monolithic$^*$} &    38.62 &    37.36 &    55.79 &    48.39 &    44.58 &    69.49 \\
			{Polyphase}      &    43.44 &    43.49 &    37.76 &    52.57 &    54.30 &    47.21 \\
			{Polyphase$^*$}  &    31.50 &    32.43 &    33.38 &    41.88 &    40.58 &    41.91 \\
			{Convolution}    &      --- &    73.79 &      --- &      --- &    83.95 &      --- \\
		\bottomrule
	\end{tabu}
	\caption{
		ALU packing percentage for AMD 6970 and AMD 5870.
	}
	\label{tab:alu-packing}
\end{table*}

In the last experiment, we were interested in a real performance.
This experiment was performed on both of the cards with CDF 9/7 wavelet.
The input as well as output raster were supplied by the texture memory.
This time, we show only the improved schemes as these always outperform the original ones.
The results are shown in \Figure{fig:evaluation-texture-improved-6970+5870}.
The horizontal axes are in a logarithmic scale, and the vertical ones express the total throughput (limited by the memory).
The {Convolution} and {Polyphase} schemes exhibit a significantly worse performance, according to the number of operations.
In contrast to this, the other schemes perform better as compared to the original separable implementation.
More specifically, the {Monolithic} and {Explosive} schemes have the very best performance.
This fact corresponds to the reduction of the number of steps (and thus the memory barriers).

The schemes presented in this paper were also subject of examination at other graphics cards under various scenarios.
Note that a link to the results is below.
Specifically, we tackled these additional cards -- NVIDIA Titan X, AMD Fury X, NVIDIA 580, and AMD 290X.
Obviously, the proposed non-separable schemes presented in this paper do not exhibit the best performance in all cases.
This is especially true for a lifting factorizations employing a longer convolution operators, as is the case of the DD 13/7 wavelet.
On the other hand, the proposed schemes seems to be the proven choice for VLIW architectures combined with a short lifting operators, e.g., the CDF 5/3 and CDF 9/7 wavelets.

This point needs to be explained in detail.
In general, the number of memory accesses, instruction dependencies, as well as barriers, decreases the ALU utilization, which then degrades the performance.
Unlike other architectures, AMD VLIW architectures pack multiple independent instructions into VLIW bundles.
Thus, amount and dependencies of instructions between each two neighboring barriers play a significant role in terms of performance.
In other words, the number of barriers in VLIW architectures plays a stronger role than in other architectures.
Indeed, on the AMD VLIW architectures, code profiling showed that memory barriers limit an average length of VLIW instructions (ALU packing percentage in \Table{tab:alu-packing}), which degrades the performance.
The ALU packing percentage refers to the percent of cores in the VLIW processor that are being utilized.
On the other (non-VLIW) architectures, the number of local memory accesses (see \Table{tab:scheme-memory}) and the number of arithmetic instructions (see \Table{tab:number-mac-barrier}) play a major role.
On such architectures, the ALU packing is not measurable due to an absence of the VLIW bundles.
For comparative purposes, the ALU packing percentage can be understood as 100\,\% for all schemes on these architectures.

\begin{table*}
	\newcommand{\1}{${}^{*}$}%
	\newcommand{\2}{${}^{**}$}%
	\centering
	\begin{tabu} to \linewidth {l|X[r]X[r]X[r]|X[r]X[r]X[r]}
			\multicolumn{1}{c}{~} & \multicolumn{3}{c}{AMD 5870} & \multicolumn{3}{c}{AMD 6970} \\
		\toprule
			scheme           &  CDF 5/3 &  CDF 9/7 &  DD 13/7 &  CDF 5/3 &  CDF 9/7 &  DD 13/7 \\
		\midrule
			{Sweldens}       &   100.00 &   100.00 &   100.00 &    95.24 &    95.24 &    95.24 \\
			{Iwahashi}       &   100.00 &   100.00 &   100.00 &    95.24 &    95.24 &    95.24 \\
			{Iwahashi$^*$}   &   100.00 &\1  83.33 &   100.00 &    95.24 &    95.24 &    95.24 \\
			{Explosive}      &   100.00 &   100.00 &   100.00 &    95.24 &    95.24 &    95.24 \\
			{Explosive$^*$}  &   100.00 &   100.00 &   100.00 &    95.24 &    95.24 &    95.24 \\
			{Monolithic}     &\1  83.33 &\1  83.33 &\1  83.33 &    95.24 &    95.24 &    95.24 \\
			{Monolithic$^*$} &\1  83.33 &\1  83.33 &\1  83.33 &    95.24 &    95.24 &    95.24 \\
			{Polyphase}      &\2  83.33 &\1  50.00 &   100.00 &    95.24 &\1  57.14 &    95.24 \\
			{Polyphase$^*$}  &   100.00 &\1  50.00 &   100.00 &    95.24 &\1  57.14 &    95.24 \\
			{Convolution}    &      --- &   100.00 &      --- &      --- &    95.24 &      --- \\
		\bottomrule
	\end{tabu}
	\caption{
		GPU occupancy measurement for AMD 6970 and AMD 5870.
		The numbers indicate a percentage.
		Explanation: \1~is limited by a local memory (LDS), \2~by registers (VGPR).
	}
	\label{tab:occupancy}
\end{table*}

It should also be interesting to show another measure provided by an OpenCL profiler.
In the first instance, consider AMD 5870 card.
Such implementations, in which threads need to store less than 5 coefficients (20 bytes) into a local memory, exhibit an occupancy 100\,\%, as can be seen in \Table{tab:occupancy}.
In detail, 256 threads in work group $\times$ 6 work groups result in occupancy 1536 of 1536 threads.
This is valid for all these implementations with the exception of {Polyphase} scheme, in which the occupancy is limited by the number of vector registers, due to an optimizing compiler.
For AMD 6970, due to the use of 256 threads in work-groups and due to maximal number 1344 of threads in multiprocesors, implementations exhibit only an occupancy 95.24\,\% (256 threads in work group $\times$ 5 work groups = 1280 of 1344).
On the other hand, considering implementations in which threads need to store less than 7 coefficients (28 bytes) into a local memory, the occupancy is not limited by a size of a local memory.

In summary, we can conclude that the reduction in lifting steps can improve performance, at least on some platforms.
This is documented by measurements in Figs.~\ref{fig:evaluation-none10-baseline-6970}, \ref{fig:evaluation-none10-baseline+improved-6970+5870}, and \ref{fig:evaluation-texture-improved-6970+5870}.
It turned out, however, that such optimization makes sense only for a short lifting operators (exemplary, degree-1 lifting filters).

For the ​sake of completeness, it should be noted that the improvement proposed in \Section{sec:improvements} can be also applied on the {Convolution}.
Doing so, the scheme achieves a slightly better performance.
However, we understand the {Convolution} scheme as the reference method.
For this reason, we leave it unimproved.
Eventually, the proposed improvement makes no sense in conjunction with the {Sweldens} scheme.

All the source codes used in this article together with all the results are available in a repository on the website of the authors' affiliation.%
\footnote{\url{http://www.fit.vutbr.cz/research/prod/?id=483}}

\section{Conclusions}
\label{sec:conclusions}

In this paper, we have proposed several non-separable lifting schemes for the calculation of the discrete wavelet transform.
The proposed schemes produce exactly the same results as the commonly used separable lifting scheme.
Using our schemes, the transform can be computed in a smaller number of steps.
On parallel architectures, this property has resulted in a smaller number of synchronizations.

Namely, we have proposed two-step \mbox{2-D} lifting scheme compatible to the commonly used four-step separable one.
Unlike the separable scheme, the proposed scheme consists of spatial predict and update operators.
Since the number of the lifting steps was halved, our scheme reduces also the number of memory barriers, which form a major bottleneck on parallel architectures.
In addition, we have proposed the three-step scheme reducing the memory access overhead.
For a moment, let $K$ denote the number of predict-update pairs.
In absolute numbers, the original separable scheme requires to write $1+4K$ coefficients per predict/update pair, whereas our three-step scheme requires $4K$ coefficients only.
Additionally, the proposed two-step scheme requires three memory cells per thread, whereas the proposed three-step scheme requires two cells only (same as the separable scheme).
Finally, we have proposed an improvement usable for all non-separable scheme, including the already known ones.
This improvement significantly reduces the number of arithmetic operations.
More specifically, the original non-separable schemes require 24 arithmetic operations for CDF 5/3 wavelet, whereas the improved variants require 18 operations only for the same case.
Even greater savings are achieved in the case of a non-factorized polyphase matrix (same as the convolution for the CDF 5/3 wavelet).
In this case, the proposed improvement reduces the number of operations from 63 to 23.
All of the proposed schemes are general and can be used in conjunction with any discrete wavelet transform.

The proposed schemes were subject to performance measurements.
In experiments on the two selected high-end GPUs (AMD Radeon HD 6970 and 5870), the proposed schemes outperform all the others for short lifting filters.
This includes the well-known CDF 5/3 and CDF 9/7 wavelets, employed, e.g., in JPEG 2000 compression standard.

Future work, we would like to do, consists of extensions to multi-dimensional systems, and extensions to another subband transforms.

\begin{acknowledgements}
This work has been supported by
the Ministry of Education, Youth and Sports of the Czech Republic from the National Programme of Sustainability (NPU II) project IT4Innovations excellence in science -- LQ1602.
\end{acknowledgements}

\bibliographystyle{spbasic_unsort}
\bibliography{sources}

\section*{Author Biographies}
\label{sec:biography}

\paragraph{\normalfont\bfseries David Barina}
received the PhD degree at the Faculty of Information Technology, Brno University of Technology, Czech Republic.
He is currently a member of the Graph@FIT group at the Department of Computer Graphics and Multimedia at FIT, Brno University of Technology.
His research interests include wavelets and fast algorithms in signal and image processing.

\paragraph{\normalfont\bfseries Michal Kula}
received the MS degree at the Faculty of Information Technology, Brno University of Technology, Czech Republic.
He is currently a PhD student and member of the Graph@FIT group at the Department of Computer Graphics and Multimedia at FIT, Brno University of Technology.

\paragraph{\normalfont\bfseries Pavel Zemcik}
received his PhD degree from the Faculty of Electrical Engineering and Computer Science, Brno University of Technology, Czech Republic.
He works as a full professor, dean and member of the Graph@FIT group at the Department of Computer Graphics and Multimedia at FIT, Brno University of Technology.
His interests include acceleration of computer vision and graphics algorithms, programmable hardware and also applications.

\end{document}